# The Benefits of Word Embeddings Features for Active Learning in Clinical Information Extraction

**Mahnoosh Kholghi, Lance De Vine, Laurianne Sitbon, Guido Zuccon**
Queensland University of Technology
`{m1.kholghi, l.devine, laurianne.sitbon, g.zuccon}@qut.edu.au`
**Anthony Nguyen**
The Australian e-Health Research Centre, CSIRO
`anthony.nguyen@csiro.au`

## Abstract

This study investigates the use of unsupervised word embeddings and sequence features for sample representation in an active learning framework built to extract clinical concepts from clinical free text. The objective is to further reduce the manual annotation effort while achieving higher effectiveness compared to a set of baseline features. Unsupervised features are derived from skip-gram word embeddings and a sequence representation approach. The comparative performance of unsupervised features and baseline handcrafted features in an active learning framework are investigated using a wide range of selection criteria including least confidence, information diversity, information density and diversity, and domain knowledge informativeness. Two clinical datasets are used for evaluation: the i2b2/VA 2010 NLP challenge and the ShARe/CLEF 2013 eHealth Evaluation Lab. Our results demonstrate significant improvements in terms of effectiveness as well as annotation effort savings across both datasets. Using unsupervised features along with baseline features for sample representation lead to further savings of up to 9% and 10% of the token and concept annotation rates, respectively.

## 1 Introduction

Active learning (AL) has recently received considerable attention in clinical information extraction, as it promises to automatically annotate clinical free text with less manual annotation effort than supervised learning approaches, while achieving the same effectiveness (Boström & Dalianis, 2012; Chen et al., 2015; Chen et al., 2012; Figueroa et al., 2012; Kholghi et al., 2015, 2016; Ohno-Machado et al., 2013). Active learning is particularly important in the clinical domain because of the costs incurred in preparing high quality annotated data as required by supervised machine learning approaches for a wide range of data analysis applications such as retrieving, reasoning, and reporting. Active learning is a human-in-the-loop process in which at each iteration, a set of informative instances is automatically selected by a *query strategy* (Settles, 2012) and annotated in order to re-train or update the supervised model (see Figure 1). The query strategy, as a key component of the AL process, plays an important role in the performance of AL approaches.

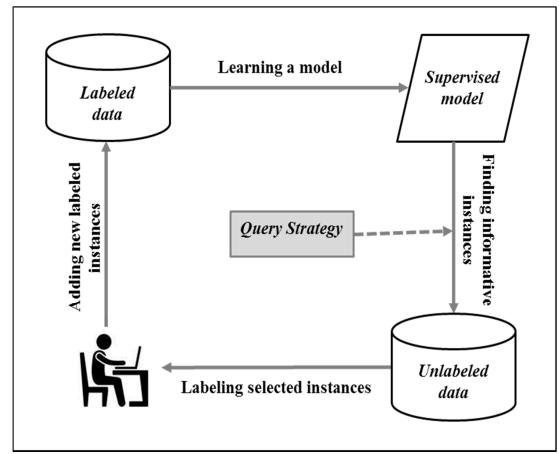

Figure 1. Active learning process.

The learning models at each iteration are typically built using supervised learning algorithms. The associated output of the learning model (i.e. the posterior probability) is usually leveraged in identifying and selecting the next set of informative instances. Hence, it is important to build ac-curate statistical models early in the process, and at each iteration. Previous studies have highlighted that the feature set, which is used to represent data instances, is an important factor that affects the stability, robustness, and effectiveness of the learning models built across the AL iterations (Kholghi et al., 2014).

In previous studies of AL for clinical information extraction, a set of common hand-crafted features, such as orthographical and morphological features, was used to build supervised models across AL iterations and suggested that more effective models would lead to reduced annotation rates in addition to improved effectiveness (Kholghi, et al., 2014, 2015, 2016). On the other hand, the application of unsupervised features, such as clustering-based representations, distributional word representations, and skip-gram word embeddings has been shown to improve fully supervised clinical information extraction systems (De Vine et al., 2015; Jonnalagadda et al., 2012; Nikfarjam et al., 2015; Tang, Wu, et al., 2013). We can therefore hypothesize that their use within an active learning framework may result in further reduction of manual annotation effort; however, no previous study has formally evaluated this in the clinical information extraction context.

In this paper, we investigate the effects of an improved sample representation using word embeddings and sequence features on an active learning framework built for clinical concept extraction. Concept extraction is a significant primary step in extracting meaningful information from clinical free text. It is a type of sequence labeling task in which sequences of terms that express meaningful concepts within a clinical setting, such as medication name, frequency, and dosage, are identified. We examine a wide range of hand-crafted and automatically generated unsupervised features to improve supervised and AL-based concept extraction systems. Our contributions are as follow:

(1) We validate the impact of word embeddings and sequence features on improving the clinical concept extraction systems, as previously studied by De Vine, et al. (2015), by using an additional dataset (ShARe/CLEF 2013 dataset) for evaluation. We generate unsupervised features using a different corpus and then investigate the combinations of features that lead to the most significant improvements on supervised models across the datasets.

(2) We demonstrate that selected combinations of unsupervised features lead to more effective models across the AL batches and also less annotation effort compared to common hand-crafted features. We do this across a selected set of query strategies.

## 2 Related Work

The two primary areas that relate to this work are: (i) the use of unsupervised sample representations in clinical information extraction, and (ii) active learning approaches for clinical information extraction.

### 2.1 Unsupervised Sample Representations in Clinical Information Extraction

The recent development of shared datasets, such as i2b2 challenges (Uzuner et al., 2010; Uzuner et al., 2011) and the ShARe/CLEF eHealth Evaluation Lab (Suominen et al., 2013) has stimulated research into new approaches to improve the current clinical information extraction systems. Unsupervised approaches to extract new features for representing data instances have proven to be key to more effective clinical information extraction systems (De Bruijn et al., 2011; De Vine, et al., 2015; Jonnalagadda, et al., 2012; Tang, Cao, et al., 2013).

Three main categories of unsupervised word representation approaches have been used in clinical information extraction systems: (1) clustering-based representations using Brown clustering (Brown et al., 1992), (2) distributional word representation using random indexing (Kanerva et al., 2000), and (3) word embeddings from neural language models, such as skip-gram word embeddings (Mikolov et al., 2013).

De Bruijn, et al. (2011) extracted clustering-based word representation features using Brown clustering and used them along with a set of hand-crafted features in developing their systems for the i2b2/VA 2010 NLP challenge. Their system achieved the highest effectiveness amongst systems in the challenge. In the same challenge, Jonnalagadda, et al. (2012) significantly improved the effectiveness of their system by adding distributional semantic features (using random indexing) to their feature set. Tang, et al. (2013) compared different word representation features extracted from Brown clustering and random indexing and found that they are complementary and when combined with common basic features the effectiveness of clinical information extraction systems increased. De Vine, et al. (2015) developed a novel approach to generate sequence level features by concatenating the accumulated and normalized word and lexical vectors of each token in a phrase or sentence. Their results demonstrated that unsupervised features generated using word embeddings and sequence level representations led to supervised models of significantly higher effectiveness compared to those built with baseline hand-crafted features.

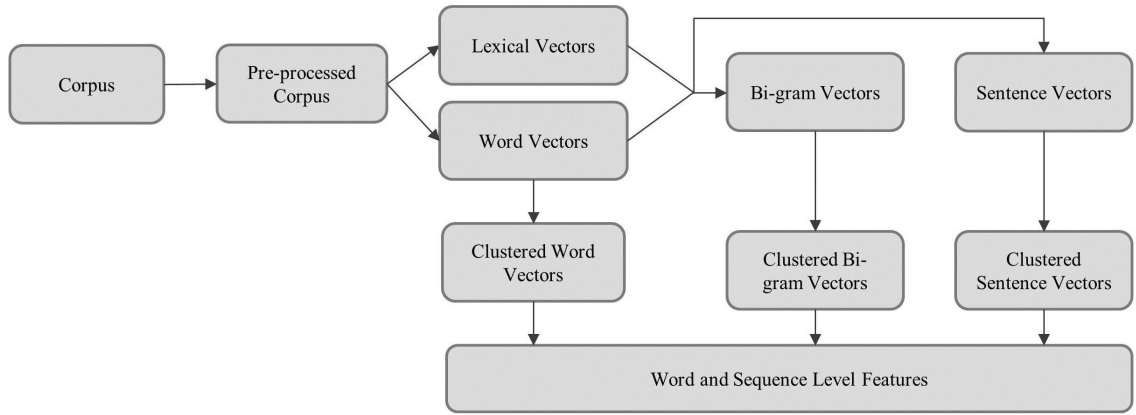

Figure 2. Word and sequence level feature generation process.

## 2.2 Active Learning in Clinical Information extraction

Active learning aims to significantly reduce the high costs of manual annotation required to build a high quality annotated data for training phase of supervised approaches.

Kholghi, et al. (2016) developed an active learning based framework to investigate the effect of AL in reducing the burden of manual annotation for clinical information extraction systems. In their framework, they apply state-of-the-art AL query strategies for sequence labelling tasks (i.e., Least Confidence (LC) and information density) to the extraction of clinical concepts. They found that AL achieves the same effectiveness as supervised learning while saving up to 77% of the total number of sequence that require manual annotation. Chen, et al. (2015) proposed new AL query strategies under groupings of uncertainty-based and diversity-based approaches. They conducted a comprehensive empirical evaluation of existing and their proposed AL approaches on the clinical concept extraction task and found that uncertainty sampling-based approaches, such as LC, resulted in a significant reduction of annotation effort compared to diversity-based approaches. Kholghi, et al. (2015) also conducted a comprehensive empirical comparison of a wide range of AL query strategies and found that the least confidence, which is an informativeness based selection criterion, is a better choice for clinical data in terms of effectiveness and annotation effort reduction. They also developed a new query strategy, called Domain Knowledge Informativeness (DKI), which makes use of external clinical resources. They showed that DKI led to a further 14% of token and concept annotation rates compared to LC.

## 3 Methodology

### 3.1 Unsupervised Sample Representation

We follow the same approach as described by De Vine, et al. (2015) to generate the unsupervised features. Figure 2 depicts our pipeline for generating the unsupervised features; these will be used to augment the supervised hand-crafted features of our classifier.

The pre-processing step includes lower-casing, substitution of matching regular expressions, and removing punctuations on the training corpus. We then generate word embeddings from the pre-processed corpus using the Skip-gram model (Mikolov, et al., 2013). We also generate lower dimensional "lexical" vectors from the pre-processed corpus, which encode character n-grams (i.e., uni-grams, bi-grams, tri-grams, tetra-grams, and skip-grams). These vectors are used to capture lexicographic patterns. A lexical vector is generated for each token by accumulating and normalizing all the n-gram vectors comprising the token.

We then use the word embeddings and the lexical vectors to construct representations for both bi-grams and sentences. First, all the lexical vectors associated with the tokens within a bi-gram or sentence are accumulated and normalized. The word embeddings for those tokens are also accumulated and normalized. Then, the resulting lexical and word vectors are concatenated and normalized to form a sequence representation for the corresponding bi-gram or sentence.

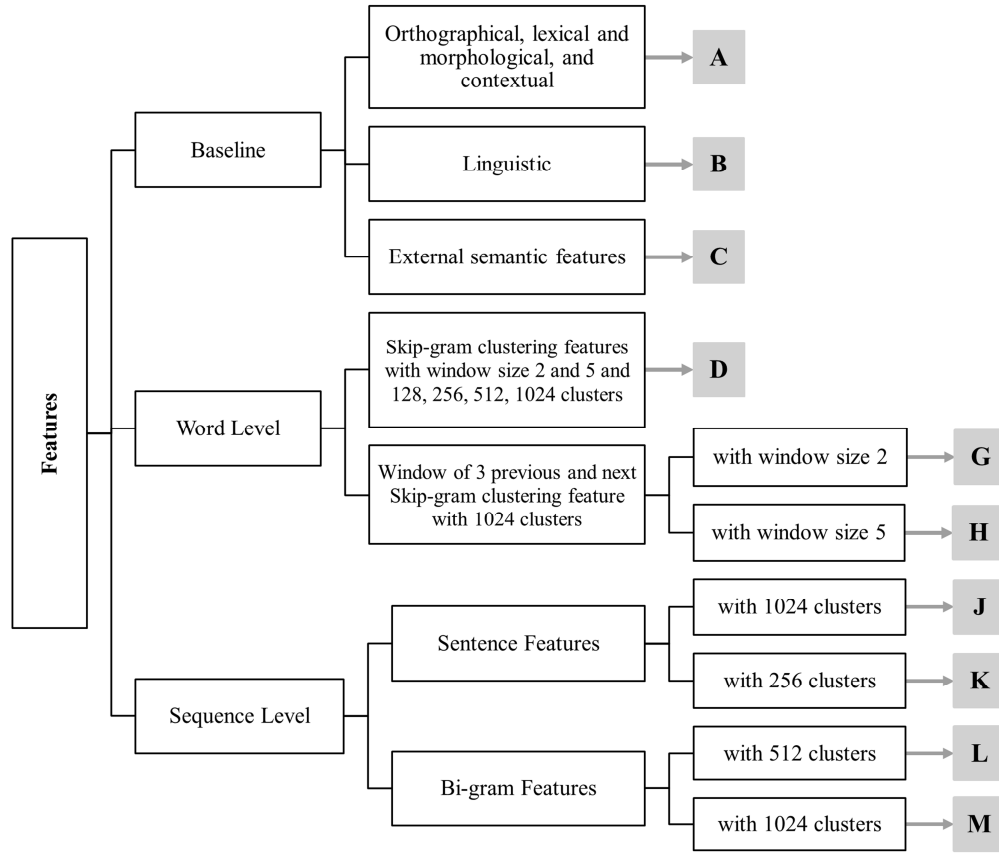

Figure 3. Description of the features used in this study.

We further cluster the word vectors, bi-gram vectors and sentence vectors to generate feature identifiers which are then used in our classifier.

### 3.2 Active Learning Query Strategies

A key element of the AL process (Figure 1) is the query strategy, which, at each iteration, selects the instances that contain the most useful information (i.e., informative samples) for the learning model. We now outline the state-of-the-art AL query strategies for clinical concept extraction (Chen, et al., 2015; Kholghi, et al., 2015).

*Least Confidence (LC)* (Culotta & McCallum, 2005) is an uncertainty-based approach in which the model's confidence (certainty) in predicting the label of a sample is the criterion to measure the informativeness of that sample. The model's confidence is estimated based on the posterior probability of the model. The less the posterior probability, the less confident the model is about the sample's label. The samples for which the model's uncertainty is the highest are the most informative for the AL model.

*Information Diversity (IDiv)* (Kholghi, et al., 2015) is based on the idea that in addition to an informativeness measure, the similarity between samples can be useful to inform the model. IDiv selects samples that are informative and diverse (i.e., those that are less similar to the labeled set).

*Information Density and Diversity (IDD)* (Kholghi, et al., 2015) is similar to IDiv with the difference that, to avoid choosing outliers, it also considers the similarity between the samples in the unlabeled set.

*Domain Knowledge Informativeness (DKI)* (Kholghi, et al., 2015) leverages the domain knowledge extracted from an external resource such as SNOMED CT, in addition to an informativeness measure, to better inform the model. The domain knowledge in DKI is estimated based on the longest span of a concept that each token belongs to, according to a pre-defined set of semantic types in the external resource.

## 4 Experimental Setup

### 4.1 Feature Groups

Figure 3 shows a short description of all the features used in this study. The baseline feature groups (A, B, C) include orthographical (regular expression patterns), lexical and morphological (suffixes/prefixes and character n-grams), contextual (window of *k* words), linguistic (POS tags (Toutanova et al., 2003)), and external semantic features (UMLS and SNOMED CT semantic groups as described in (Kholghi, et al., 2015)).

As the previous work demonstrated, learning word embeddings and sequence features from a clinical corpus with an adequate amount of data, and a good coverage of the target data, results in higher effectiveness compared to a general or relatively small clinical corpus (De Vine, et al., 2015). In this study, we use a clinical corpus composed of the concatenation of the i2b2/VA 2010 train set (Uzuner, et al., 2011), the MedTrack collection (Voorhees & Tong, 2011), and the ShARe/CLEF 2013 train set (Suominen, et al., 2013) to generate word embeddings.

### 4.2 Supervised and Active Learning Settings

In this study, we use an incremental, pool-based, active learning framework (Kholghi, et al., 2014, 2016). We build models across AL batches using tuned linear chain Conditional Random Fields (CRFs) (Kholghi, et al., 2014; Lafferty et al., 2001) with different feature groups. The implementation of CRFs for both supervised and active learning is based on the MALLET toolkit (McCallum, 2002). In this study, Random Sampling (RS) is used as a baseline for the AL framework. RS randomly selects samples at each iteration. All active learning and random sampling baseline setups including the initial labeled set and batch size (i.e., both less than 1% of the size of the train set) are based on previous findings (Kholghi, et al., 2015, 2016).

### 4.3 Datasets

We use the annotated train sets developed for the concept extraction task in the i2b2/VA 2010 NLP challenge (Uzuner, et al., 2011) and ShARe/CLEF 2013 eHealth Evaluation Lab (Task 1) (Pradhan et al., 2013) to build learning models across AL batches using different feature groups.

Table 1. Number of documents (#doc) and sequences (#seq) in the train and test sets of the two considered datasets.

| | Train Set | | Test Set | |
|---|---|---|---|---|
| | #doc | #seq | #doc | #seq |
| **i2b2/VA 2010** | 349 | 30,673 | 477 | 45,025 |
| **ShARe/CLEF 2013** | 200 | 10,171 | 100 | 9,273 |

The corresponding test set of each dataset is used to evaluate the effect of feature groups on the performance of models built across AL batches (see Table 1). The i2b2/VA 2010 task comprises the extraction of clinical problems, tests and treatments from clinical reports, while the ShARe/CLEF 2013 eHealth Evaluation Lab (task 1) requires to identify mentions of disorders.

### 4.4 Evaluation measures

In our evaluation, the learning model effectiveness is measured by Precision, Recall and F1-measure. The evaluation measures are computed on the test set using MALLET's multi-segmentation evaluator (McCallum, 2002). To demonstrate statistically significant improvements on F1- measures, we perform a 5*2 cross validated paired t-test (Dietterich, 1998).

The performance of the AL framework is evaluated using Annotation Rate (AR), which measures the number of Sequences (SAR), Tokens (TAR), and Concepts (CAR) required by the AL framework to reach the target supervised effectiveness. The lower the annotation rate, the better the AL framework is considered to be.

$$AR = \frac{\text{\# labeled annotation units used by AL}}{\text{\# total labeled annotation units in train set}}$$

## 5 Results

### 5.1 Target Supervised Performance

Table 2 presents the effectiveness of the supervised CRFs models, which employ all the labeled instances in the train sets of the considered datasets, using the different combinations of features described in Figure 3. The highest effectiveness obtained in each feature group is highlighted in bold.

Table 2 shows that the inclusion of the unsupervised word and sequence level features improves the effectiveness of the supervised model compared to the best baseline feature set ABC. The models' effectiveness built using feature groups ABCD, ABCDGH, ABCDGHK, and ABCDGHJKM are selected for subsequent active learning experiments as target supervised effectiveness, because they result in considerable improvements in the supervised models' effectiveness across both datasets.

Table 2. Supervised target performance for all sets of features. Statistically significant improvements
(p<0.05) for F1 when compared with ABC are indicated by *.

| | Features | i2b2/VA 2010 | | | ShARe/CLEF 2013 | | |
|---|---|---|---|---|---|---|---|
| | | Precision | Recall | F1 measure | Precision | Recall | F1 measure |
| Baseline | Word | 0.6571 | 0.6011 | 0.6279 | 0.2225 | 0.4317 | 0.2936 |
| | A | 0.8404 | 0.8031 | 0.8213 | 0.7858 | 0.6461 | 0.7091 |
| | B | 0.6167 | 0.6006 | 0.6085 | 0.5157 | 0.4027 | 0.4523 |
| | C | 0.7691 | 0.6726 | 0.7192 | 0.7022 | 0.5118 | 0.5921 |
| | BC | 0.7269 | 0.712 | 0.7194 | 0.7163 | 0.518 | 0.6012 |
| | AB | 0.8368 | 0.8038 | 0.82 | 0.7832 | 0.6472 | 0.7087 |
| | AC | 0.8378 | 0.8059 | 0.8216 | 0.8035 | 0.6808 | 0.7371 |
| | **ABC** | 0.8409 | 0.8066 | **0.8234** | 0.8095 | 0.6804 | **0.7394** |
| Word Level | D | 0.7773 | 0.7393 | 0.7578 | 0.6815 | 0.5581 | 0.6137 |
| | GH | 0.8056 | 0.7547 | 0.7793 | 0.7225 | 0.5625 | 0.6325 |
| | ABCD | 0.8424 | 0.8127 | 0.8273 | 0.8042 | 0.6916 | 0.7436 |
| | **ABCDGH** | 0.8502 | 0.8124 | **0.8309*** | 0.8092 | 0.6898 | **0.7448*** |
| Sequence Level | J | 0.6551 | 0.6242 | 0.6393 | 0.6564 | 0.4054 | 0.5012 |
| | K | 0.6852 | 0.6433 | 0.6636 | 0.6305 | 0.4189 | 0.5033 |
| | ABCDGHJ | 0.8488 | 0.8126 | 0.8303* | 0.7992 | 0.6916 | 0.7415 |
| | **ABCDGHK** | 0.8495 | 0.8132 | **0.8309*** | 0.8111 | 0.69 | **0.7457*** |
| | ABCDGHJK | 0.8449 | 0.8116 | 0.8279 | 0.8093 | 0.6889 | 0.7443 |
| | L | 0.7361 | 0.6169 | 0.6713 | 0.7015 | 0.3854 | 0.4975 |
| | M | 0.7531 | 0.6358 | 0.6895 | 0.672 | 0.3924 | 0.4955 |
| | ABCDGHJKL | 0.8458 | 0.8086 | 0.8268 | 0.8068 | 0.6881 | 0.7427 |
| | **ABCDGHJKM** | 0.8488 | 0.8113 | **0.8296*** | 0.8105 | 0.69 | **0.7454*** |
| | ABCDGHJKLM | 0.8447 | 0.8062 | 0.825 | 0.8117 | 0.6873 | 0.7444 |

### 5.2 Active Learning Performance

We now consider the performance of the active learning framework in terms of annotation rates.

It is important to note that in these experiments, the models built across AL batches, using selected feature sets, are required to reach the target supervised effectiveness achieved using the corresponding feature set (F1-measures in Table 2).

Table 3 presents SAR, TAR and CAR for different AL query strategies and for the Random Sampling baseline. The most effective feature sets, compared to the baseline feature set ABC (highlighted in gray), for the models built across AL batches using different query strategies are highlighted in bold.

Word and sequence representations result in less annotation effort across all query strategies in both datasets compared to the hand-crafted feature set. We observe 9% and 10% reduction in token (TAR) and concept (CAR) annotation rates for the IDiv query strategy (highlighted in orange) when using ABCDGH feature set compared to the baseline ABC feature set in ShARe/CLEF 2013 dataset. The same feature set (ABCDGH) results in 4% and 6% less TAR and CAR in i2b2/VA 2010 dataset (highlighted in green) compared to the baseline ABC feature set when using LC as the query strategy.

Generally, the addition of word level features (D, G, and H) gives the best results. Also, on occasions, the addition of sequence level features (J, K, and M) gives further improvements, although not consistently. The previous study also showed that the addition of sequence level features results in less remarkable improvement on supervised models' effectiveness compared to word level features (De Vine, et al., 2015).

### 6 Discussion

The results from our empirical evaluation confirm the previous findings suggesting that the use of unsupervised features significantly improves clinical information extraction systems in a supervised learning setting (De Vine, et al., 2015). Here we have further studied the use of these features within an active learning framework.

Our results highlight that the use of unsupervised word and sequence level features not only increases the effectiveness of the models built across AL batches, but also leads to lower manual annotation efforts in the active learning framework compared to the baseline feature set ABC (no unsupervised features). We can assume that the reason is that the better the sample representation, the stronger the updated model is in terms of effectiveness at each iteration of active learning. This means that AL query strategies use a better updated model at each iteration and therefore choose a better set of informative instances. Hence, by using these data representations, AL requires a smaller number of sequences, tokens, and concepts to reach the target supervised effectiveness. This, in turn, translates into lower annotation rates. However, not all combinations of different features always lead to lower annotation rates in the AL framework (Kholghi, et al., 2014).

Table 3. Annotation rates for all active learning query strategies and the baseline RS using different sample representations (feature groups). Results for the baseline feature set (ABC) are highlighted in gray.

| Query Strategy | Features | i2b2/VA 2010 | | | ShARe/CLEF 2013 | | |
|---|---|---|---|---|---|---|---|
| | | SAR | TAR | CAR | SAR | TAR | CAR |
| RS | ABC | 90% | 90% | 90% | 97% | 97% | 98% |
| | ABCD | 88% | 88% | 88% | 83% | 84% | 83% |
| | ABCDGH | **82%** | **82%** | **82%** | 88% | 88% | 87% |
| | ABCDGHK | 88% | 88% | 88% | 91% | 91% | 91% |
| | ABCDGHJKM | 87% | 87% | 87% | **76%** | **76%** | **76%** |
| LC | ABC | 24% | 43% | 55% | 24% | 38% | 63% |
| | ABCD | 22% | 40% | 50% | **19%** | 31% | 55% |
| | ABCDGH | **20%** | **39%** | **49%** | 20% | 33% | 58% |
| | ABCDGHK | **20%** | 39% | 49% | 22% | 35% | 61% |
| | ABCDGHJKM | 22% | 41% | 52% | **20%** | **33%** | **57%** |
| IDiv | ABC | 23% | 41% | 52% | 24% | 38% | 64% |
| | ABCD | **20%** | 37% | 48% | 20% | 31% | 55% |
| | ABCDGH | **20%** | 39% | 50% | **18%** | **29%** | **54%** |
| | ABCDGHK | 22% | 42% | 52% | 20% | 31% | 57% |
| | ABCDGHJKM | 22% | 41% | 52% | 20% | 31% | 56% |
| IDD | ABC | 22% | 41% | 52% | 25% | 41% | 66% |
| | ABCD | 22% | 41% | 51% | 23% | 38% | 62% |
| | ABCDGH | **20%** | **39%** | **49%** | **20%** | **33%** | **57%** |
| | ABCDGHK | 22% | 40% | 51% | 21% | 35% | 59% |
| | ABCDGHJKM | **20%** | **39%** | **49%** | 22% | 36% | 61% |
| DKI | ABC | 22% | 27% | 37% | 20% | 31% | 57% |
| | ABCD | 19% | 25% | 36% | **17%** | **27%** | **52%** |
| | ABCDGH | **18%** | **24%** | **35%** | 18% | 29% | 54% |
| | ABCDGHK | **18%** | **24%** | **35%** | 20% | 30% | 55% |
| | ABCDGHJKM | 19% | 25% | 36% | **19%** | 28% | 53% |

We thus next study the trade-off between effectiveness (F1 measure from Table 2) and annotation rate (CAR from Table 3) to better understand the performance of five selected feature groups (ABC, ABCD, ABCDGH, and ABCDGHK). Figure 4 demonstrates the concept annotation rate (CAR) values (horizontal axis) for the best performing query strategy, in each dataset, when reaching: (1) the corresponding target supervised effectiveness for each feature set showed by ■, and (2) a fixed effectiveness for all feature sets showed by ◆. These values are depicted against the effectiveness when training on the full train set (vertical axis) for each feature set (F1 measure from Table 2). We are presenting these for LC and IDiv for i2b2/VA 2010 and ShARe/CLEF datasets, respectively as they achieved the lowest concept annotation rates as discussed in section 5.2. The fixed effectiveness for all feature sets is determined as follow: F1 measure = 0.80 for i2b2/VA 2010 and F1 measure = 0.70 for ShARe/CLEF 2013. The aim of this analysis is to verify whether improvements in terms of supervised effectiveness when using different feature sets (F1 measure from Table 2) necessarily scale into improvements in CAR (i.e., lower annotation effort) and whether the same behavior is observed in terms of annotation effort reduction when a fixed F1 measure value is considered for all feature groups. It is important to note that the higher the F1 measure and the lower the CAR, the better the feature set. Hence, those points towards the left upper corner of both plots in Figure 4 perform better both in terms of effectiveness and annotation rate. Points marked with the same symbol should be compared to each other.

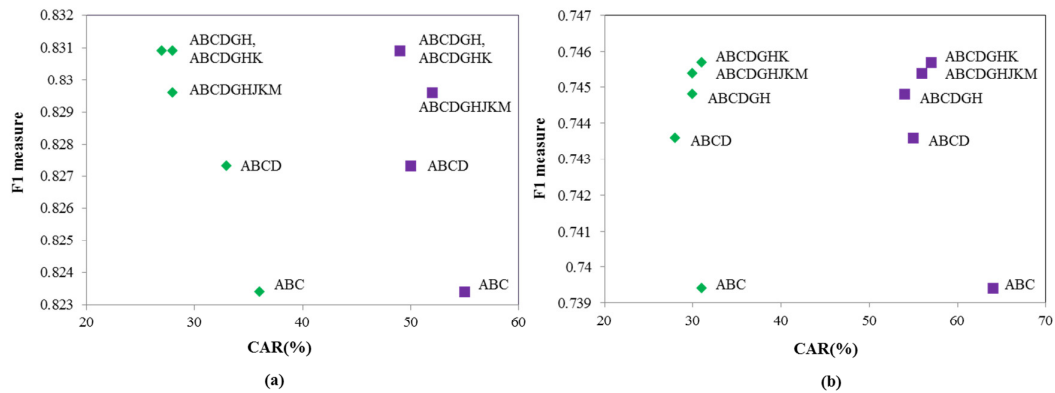

Figure 4. Analysis of concept annotation rates (CAR) (horizontal axis) at (1) target supervised effectiveness for each feature set (■), and (2) a fixed effectiveness for all feature sets (♦) with respect to the corresponding F1 measure for each feature set from Table 2 (vertical axis). (a) i2b2/VA 2010; (b) ShARe/CLEF 2013.

In terms of target supervised effectiveness (■), Figure 4 shows that feature groups ABCDGH and ABCDGHK outperform the other feature groups in i2b2/VA 2010 dataset, both in terms of effectiveness (F1 measure) and annotation rate (CAR). While ABCDGH achieves the best CAR (i.e., the lowest) in ShARe/CLEF 2013 dataset, it is not the best performing feature group in terms of supervised effectiveness. The highest F1 measure was achieved by feature group ABCDGHK in this dataset. The same pattern is observed when considering a fixed F1 measure value (♦). Hence, the feature set that leads to a supervised model with the highest effectiveness (F1 measure) does not always lead to an AL model with the lowest annotation rate. These results demonstrate that improving supervised models built across the AL batches does not necessarily guarantee a reduction in annotation rates.

Interestingly, the updated model has no role in selecting the next batch of instances when using the Random Sampling baseline, as this randomly selects instances at each iteration. Yet, a better feature set (e.g., ABCDGHJKM) helps RS to reduce the annotation rate. If we compare the updated models at the same batch of RS using different data representations, for instance ABC vs. ABCDGHJKM, we observe that by even adding random instances to the labeled set, more information is injected into the updated model using the feature set ABCDGHJKM compared to ABC. This suggests that RS with unsupervised features has a reduced rate of annotation errors compared to using the ABC feature set.

These results can be summarized into the following observations:

- A better sample representation using unsupervised features leads to higher effectiveness and less manual annotation effort not only in an AL framework, but also in a Random Sampling approach.
- Although there is a relationship between high effectiveness and low annotation effort, not all combinations of features conducive to the highest effectiveness necessarily lead to the lowest annotation effort.
- The combination of word level features (D, G, and H) with the baseline handcrafted features, i.e., ABCDGH, generally performs better than the other feature combinations across all AL query strategies and datasets, both in terms of effectiveness and annotation rates.

## 7 Conclusion

This paper presented an analysis of different data representations using a wide range of feature sets and investigated their impact on active learning performance in terms of both model effectiveness and annotation effort reduction. We believe this is the first study analyzing the effect of unsupervised sample representation using word embeddings and sequence level features on an active learning framework built for clinical information extraction.

The empirical results highlighted the benefits of unsupervised features in achieving higher effectiveness and lower manual annotation effort in our AL framework. Word and sequence level features significantly increase the effectiveness of the models built across AL batches. In addition, compared to the baseline feature set, they reduce the manual annotation effort by using a small number of sequences, tokens, and concepts to reach the target supervised performance. Hence, it can be concluded that the manual annotation of clinical free text for information extraction applications can be accelerated using an improved sample representation in an active learning framework. While this could seem intuitive, we have also shown that improvements demonstrated in a fully supervised framework do not necessarily translate into improvements in an active learning framework.